# Dynamic Backtracking

**Matthew L. Ginsberg**                                    GINSBERG@CS.UOREGON.EDU
*CIRL, University of Oregon,*
*Eugene, OR 97403-1269 USA*

## Abstract

Because of their occasional need to return to shallow points in a search tree, existing backtracking methods can sometimes erase meaningful progress toward solving a search problem. In this paper, we present a method by which backtrack points can be moved deeper in the search space, thereby avoiding this difficulty. The technique developed is a variant of dependency-directed backtracking that uses only polynomial space while still providing useful control information and retaining the completeness guarantees provided by earlier approaches.

## 1. Introduction

Imagine that you are trying to solve some constraint-satisfaction problem, or CSP. In the interests of definiteness, I will suppose that the CSP in question involves coloring a map of the United States subject to the restriction that adjacent states be colored differently.

Imagine we begin by coloring the states along the Mississippi, thereby splitting the remaining problem in two. We now begin to color the states in the western half of the country, coloring perhaps half a dozen of them before deciding that we are likely to be able to color the rest. Suppose also that the last state colored was Arizona.

At this point, we change our focus to the eastern half of the country. After all, if we can't color the eastern half because of our coloring choices for the states along the Mississippi, there is no point in wasting time completing the coloring of the western states.

We successfully color the eastern states and then return to the west. Unfortunately, we color New Mexico and Utah and then get stuck, unable to color (say) Nevada. What's more, backtracking doesn't help, at least in the sense that changing the colors for New Mexico and Utah alone does not allow us to proceed farther. Depth-first search would now have us backtrack to the eastern states, trying a new color for (say) New York in the vain hope that this would solve our problems out West.

This is obviously pointless; the blockade along the Mississippi makes it impossible for New York to have any impact on our attempt to color Nevada or other western states. What's more, we are likely to examine every *possible* coloring of the eastern states before addressing the problem that is actually the source of our difficulties.

The solutions that have been proposed to this involve finding ways to backtrack directly to some state that might actually allow us to make progress, in this case Arizona or earlier. Dependency-directed backtracking (Stallman & Sussman, 1977) involves a direct backtrack to the source of the difficulty; backjumping (Gaschnig, 1979) avoids the computational overhead of this technique by using syntactic methods to estimate the point to which backtrack is necessary.

In both cases, however, note that although we backtrack to the source of the problem, we backtrack *over* our successful solution to half of the original problem, discarding our solution to the problem of coloring the states in the East. And once again, the problem is worse than this – after we recolor Arizona, we are in danger of solving the East yet again before realizing that our new choice for Arizona needs to be changed after all. We won't examine every possible coloring of the eastern states, but we are in danger of rediscovering our successful coloring an exponential number of times.

This hardly seems sensible; a human problem solver working on this problem would simply ignore the East if possible, returning directly to Arizona and proceeding. Only if the states along the Mississippi needed new colors would the East be reconsidered – and even then only if no new coloring could be found for the Mississippi that was consistent with the eastern solution.

In this paper we formalize this technique, presenting a modification to conventional search techniques that is capable of backtracking not only to the most recently expanded node, but also directly to a node elsewhere in the search tree. Because of the dynamic way in which the search is structured, we refer to this technique as *dynamic backtracking*.

A more specific outline is as follows: We begin in the next section by introducing a variety of notational conventions that allow us to cast both existing work and our new ideas in a uniform computational setting. Section 3 discusses backjumping, an intermediate between simple chronological backtracking and our ideas, which are themselves presented in Section 4. An example of the dynamic backtracking algorithm in use appears in Section 5 and an experimental analysis of the technique in Section 6. A summary of our results and suggestions for future work are in Section 7. All proofs have been deferred to an appendix in the interests of continuity of exposition.

## 2. Preliminaries

**Definition 2.1** *By a* constraint satisfaction problem $(I, V, \kappa)$ *we will mean a set $I$ of variables; for each $i \in I$, there is a set $V_i$ of possible values for the variable $i$. $\kappa$ is a set of constraints, each a pair $(J, P)$ where $J = (j_1, \ldots, j_k)$ is an ordered subset of $I$ and $P$ is a subset of $V_{j_1} \times \cdots \times V_{j_k}$.*

*A* solution *to the* CSP *is a set $v_i$ of values for each of the variables in $I$ such that $v_i \in V_i$ for each $i$ and for every constraint $(J, P)$ of the above form in $\kappa$, $(v_{j_1}, \ldots, v_{j_k}) \in P$.*

In the example of the introduction, $I$ is the set of states and $V_i$ is the set of possible colors for the state $i$. For each constraint, the first part of the constraint is a pair of adjacent states and the second part is a set of allowable color combinations for these states.

Our basic plan in this paper is to present formal versions of the search algorithms described in the introduction, beginning with simple depth-first search and proceeding to backjumping and dynamic backtracking. As a start, we make the following definition of a partial solution to a CSP:

**Definition 2.2** *Let $(I, V, \kappa)$ be a* CSP. *By a* partial solution *to the* CSP *we mean an ordered subset $J \subseteq I$ and an assignment of a value to each variable in $J$.*

*We will denote a partial solution by a tuple of ordered pairs, where each ordered pair $(i, v)$ assigns the value $v$ to the variable $i$. For a partial solution $P$, we will denote by $\overline{P}$ the set of variables assigned values by $P$.*

Constraint-satisfaction problems are solved in practice by taking partial solutions and extending them by assigning values to new variables. In general, of course, not any value can be assigned to a variable because some are inconsistent with the constraints. We therefore make the following definition:

**Definition 2.3** *Given a partial solution $P$ to a CSP, an* eliminating explanation *for a variable $i$ is a pair $(v, S)$ where $v \in V_i$ and $S \subseteq \overline{P}$. The intended meaning is that $i$ cannot take the value $v$ because of the values already assigned by $P$ to the variables in $S$. An* elimination mechanism *$\epsilon$ for a CSP is a function that accepts as arguments a partial solution $P$, and a variable $i \notin \overline{P}$. The function returns a (possibly empty) set $\epsilon(P, i)$ of eliminating explanations for $i$.*

For a set $E$ of eliminating explanations, we will denote by $\widehat{E}$ the values that have been identified as eliminated, ignoring the reasons given. We therefore denote by $\widehat{\epsilon}(P, i)$ the set of values eliminated by elements of $\epsilon(P, i)$.

Note that the above definition is somewhat flexible with regard to the amount of work done by the elimination mechanism – all values that violate completed constraints might be eliminated, or some amount of lookahead might be done. We will, however, make the following assumptions about all elimination mechanisms:

1. They are *correct*. For a partial solution $P$, if the value $v_i \notin \widehat{\epsilon}(P, i)$, then every constraint $(S, T)$ in $\kappa$ with $S \subseteq \overline{P} \cup \{i\}$ is satisfied by the values in the partial solution and the value $v_i$ for $i$. These are the constraints that are complete after the value $v_i$ is assigned to $i$.

2. They are *complete*. Suppose that $P$ is a partial solution to a CSP, and there is some solution that extends $P$ while assigning the value $v$ to $i$. If $P'$ is an extension of $P$ with $(v, E) \in \epsilon(P', i)$, then

$$E \cap \left(\overline{P'} - \overline{P}\right) \neq \emptyset \tag{1}$$

   In other words, whenever $P$ can be successfully extended after assigning $v$ to $i$ but $P'$ cannot be, at least one element of $P' - P$ is identified as a possible reason for the problem.

3. They are *concise*. For a partial solution $P$, variable $i$ and eliminated value $v$, there is at most a single element of the form $(v, E) \in \epsilon(P, i)$. Only one reason is given why the variable $i$ cannot have the value $v$.

**Lemma 2.4** *Let $\epsilon$ be a complete elimination mechanism for a CSP, let $P$ be a partial solution to this CSP and let $i \notin \overline{P}$. Now if $P$ can be successfully extended to a complete solution after assigning $i$ the value $v$, then $v \notin \widehat{\epsilon}(P, i)$.*

I apologize for the swarm of definitions, but they allow us to give a clean description of depth-first search:

**Algorithm 2.5 (Depth-first search)** *Given as inputs a constraint-satisfaction problem and an elimination mechanism $\epsilon$:*

1. *Set $P = \emptyset$. $P$ is a partial solution to the CSP. Set $E_i = \emptyset$ for each $i \in I$; $E_i$ is the set of values that have been eliminated for the variable $i$.*

2. *If $\overline{P} = I$, so that $P$ assigns a value to every element in $I$, it is a solution to the original problem. Return it. Otherwise, select a variable $i \in I - \overline{P}$. Set $E_i = \widehat{\epsilon}(P, i)$, the values that have been eliminated as possible choices for $i$.*

3. *Set $S = V_i - E_i$, the set of remaining possibilities for $i$. If $S$ is nonempty, choose an element $v \in S$. Add $(i, v)$ to $P$, thereby setting $i$'s value to $v$, and return to step 2.*

4. *If $S$ is empty, let $(j, v_j)$ be the last entry in $P$; if there is no such entry, return failure. Remove $(j, v_j)$ from $P$, add $v_j$ to $E_j$, set $i = j$ and return to step 3.*

We have written the algorithm so that it returns a single answer to the CSP; the modification to accumulate all such answers is straightforward.

The problem with Algorithm 2.5 is that it looks very little like conventional depth-first search, since instead of recording the unexpanded children of any particular node, we are keeping track of the *failed siblings* of that node. But we have the following:

**Lemma 2.6** *At any point in the execution of Algorithm 2.5, if the last element of the partial solution $P$ assigns a value to the variable $i$, then the unexplored siblings of the current node are those that assign to $i$ the values in $V_i - E_i$.*

**Proposition 2.7** *Algorithm 2.5 is equivalent to depth-first search and therefore complete.*

As we have remarked, the basic difference between Algorithm 2.5 and a more conventional description of depth-first search is the inclusion of the elimination sets $E_i$. The conventional description expects nodes to include pointers back to their parents; the siblings of a given node are found by examining the children of that node's parent. Since we will be reorganizing the space as we search, this is impractical in our framework.

It might seem that a more natural solution to this difficulty would be to record not the values that have been *eliminated* for a variable $i$, but those that remain to be considered. The technical reason that we have not done this is that it is much easier to maintain elimination information as the search progresses. To understand this at an intuitive level, note that when the search backtracks, the conclusion that has implicitly been drawn is that a particular node fails to expand to a solution, as opposed to a conclusion about the currently unexplored portion of the search space. It should be little surprise that the most efficient way to manipulate this information is by recording it in approximately this form.

## 3. Backjumping

How are we to describe dependency-directed backtracking or backjumping in this setting? In these cases, we have a partial solution and have been forced to backtrack; these more sophisticated backtracking mechanisms use information about the *reason* for the failure to identify backtrack points that might allow the problem to be addressed. As a start, we need to modify Algorithm 2.5 to maintain the explanations for the eliminated values:

**Algorithm 3.1** *Given as inputs a constraint-satisfaction problem and an elimination mechanism $\epsilon$:*

1. *Set $P = E_i = \emptyset$ for each $i \in I$. $E_i$ is a set of eliminating explanations for $i$.*

2. *If $\overline{P} = I$, return $P$. Otherwise, select a variable $i \in I - \overline{P}$. Set $E_i = \epsilon(P, i)$.*

3. *Set $S = V_i - \widehat{E}_i$. If $S$ is nonempty, choose an element $v \in S$. Add $(i, v)$ to $P$ and return to step 2.*

4. *If $S$ is empty, let $(j, v_j)$ be the last entry in $P$; if there is no such entry, return failure. Remove $(j, v_j)$ from $P$. We must have $\widehat{E}_i = V_i$, so that every value for $i$ has been eliminated; let $E$ be the set of all variables appearing in the explanations for each eliminated value. Add $(v_j, E - \{j\})$ to $E_j$, set $i = j$ and return to step 3.*

**Lemma 3.2** *Let $P$ be a partial solution obtained during the execution of Algorithm 3.1, and let $i \in \overline{P}$ be a variable assigned a value by $P$. Now if $P' \subseteq P$ can be successfully extended to a complete solution after assigning $i$ the value $v$ but $(v, E) \in E_i$, we must have*

$$E \cap (\overline{P} - \overline{P'}) \neq \emptyset$$

In other words, the assignment of a value to some variable in $\overline{P} - \overline{P'}$ is correctly identified as the source of the problem.

Note that in step 4 of the algorithm, we could have added $(v_j, E \cap \overline{P})$ instead of $(v_j, E - \{j\})$ to $E_j$; either way, the idea is to remove from $E$ any variables that are no longer assigned values by $P$.

In backjumping, we now simply change our backtrack method; instead of removing a single entry from $P$ and returning to the variable assigned a value prior to the problematic variable $i$, we return to a variable that has actually had an impact on $i$. In other words, we return to some variable in the set $E$.

**Algorithm 3.3 (Backjumping)** *Given as inputs a constraint-satisfaction problem and an elimination mechanism $\epsilon$:*

1. *Set $P = E_i = \emptyset$ for each $i \in I$.*

2. *If $\overline{P} = I$, return $P$. Otherwise, select a variable $i \in I - \overline{P}$. Set $E_i = \epsilon(P, i)$.*

3. *Set $S = V_i - \widehat{E}_i$. If $S$ is nonempty, choose an element $v \in S$. Add $(i, v)$ to $P$ and return to step 2.*

4. *If $S$ is empty, we must have $\widehat{E}_i = V_i$. Let $E$ be the set of all variables appearing in the explanations for each eliminated value.*

5. *If $E = \emptyset$, return failure. Otherwise, let $(j, v_j)$ be the last entry in $P$ such that $j \in E$. Remove from $P$ this entry and any entry following it. Add $(v_j, E \cap \overline{P})$ to $E_j$, set $i = j$ and return to step 3.*

In step 5, we add $(v_j, E \cap \overline{P})$ to $E_j$, removing from $E$ any variables that are no longer assigned values by $P$.

**Proposition 3.4** *Backjumping is complete and always expands fewer nodes than does depth-first search.*

Let us have a look at this in our map-coloring example. If we have a partial coloring $P$ and are looking at a specific state $i$, suppose that we denote by $C$ the set of colors that are obviously illegal for $i$ because they conflict with a color already assigned to one of $i$'s neighbors.

One possible elimination mechanism returns as $\epsilon(P, i)$ a list of $(c, \overline{P})$ for each color $c \in C$ that has been used to color a neighbor of $i$. This reproduces depth-first search, since we gradually try all possible colors but have no idea what went wrong when we need to backtrack since every colored state is included in $\overline{P}$. A far more sensible choice would take $\epsilon(P, i)$ to be a list of $(c, \{n\})$ where $n$ is a neighbor that is already colored $c$. This would ensure that we backjump to a neighbor of $i$ if no coloring for $i$ can be found.

If this causes us to backjump to another state $j$, we will add $i$'s neighbors to the eliminating explanation for $j$'s original color, so that if we need to backtrack still further, we consider neighbors of either $i$ or $j$. This is as it should be, since changing the color of one of $i$'s other neighbors might allow us to solve the coloring problem by reverting to our original choice of color for the state $j$.

We also have:

**Proposition 3.5** *The amount of space needed by backjumping is $o(i^2 v)$, where $i = |I|$ is the number of variables in the problem and $v$ is the number of values for that variable with the largest value set $V_i$.*

This result contrasts sharply with an approach to CSPs that relies on truth-maintenance techniques to maintain a list of nogoods (de Kleer, 1986). There, the number of nogoods found can grow linearly with the time taken for the analysis, and this will typically be exponential in the size of the problem. Backjumping avoids this problem by resetting the set $E_i$ of eliminating explanations in step 2 of Algorithm 3.3.

The description that we have given is quite similar to that developed in (Bruynooghe, 1981). The explanations there are somewhat coarser than ours, listing all of the variables that have been involved in *any* eliminating explanation for a particular variable in the CSP, but the idea is essentially the same. Bruynooghe's eliminating explanations can be stored in $o(i^2)$ space (instead of $o(i^2 v)$), but the associated loss of information makes the technique less effective in practice. This earlier work is also a description of backjumping only, since intermediate information is erased as the search proceeds.

## 4. Dynamic backtracking

We finally turn to new results. The basic problem with Algorithm 3.3 is not that it backjumps to the wrong place, but that it needlessly erases a great deal of the work that has been done thus far. At the very least, we can retain the values selected for variables that are backjumped over, in some sense moving the backjump variable to the end of the partial solution in order to replace its value without modifying the values of the variables that followed it.

There is an additional modification that will probably be clearest if we return to the example of the introduction. Suppose that in this example, we color only *some* of the eastern states before returning to the western half of the country. We reorder the variables in order to backtrack to Arizona and eventually succeed in coloring the West without disturbing the colors used in the East.

Unfortunately, when we return East backtracking is required and we find ourselves needing to change the coloring on some of the eastern states with which we dealt earlier. The ideas that we have presented will allow us to avoid erasing our solution to the problems out West, but if the search through the eastern states is to be efficient, we will need to retain the information we have about the portion of the East's search space that has been eliminated. After all, if we have determined that New York cannot be colored yellow, our changes in the West will not reverse this conclusion – the Mississippi really does isolate one section of the country from the other.

The machinery needed to capture this sort of reasoning is already in place. When we backjump over a variable $k$, we should retain not only the choice of value for $k$, but also $k$'s elimination set. We do, however, need to remove from this elimination set any entry that involves the eventual backtrack variable $j$, since these entries are no longer valid – they depend on the assumption that $j$ takes its old value, and this assumption is now false.

**Algorithm 4.1 (Dynamic backtracking I)** *Given as inputs a constraint-satisfaction problem and an elimination mechanism $\epsilon$:*

1. *Set $P = E_i = \emptyset$ for each $i \in I$.*

2. *If $\overline{P} = I$, return $P$. Otherwise, select a variable $i \in I - \overline{P}$. Set $E_i = E_i \cup \epsilon(P, i)$.*

3. *Set $S = V_i - \widehat{E}_i$. If $S$ is nonempty, choose an element $v \in S$. Add $(i, v)$ to $P$ and return to step 2.*

4. *If $S$ is empty, we must have $\widehat{E}_i = V_i$; let $E$ be the set of all variables appearing in the explanations for each eliminated value.*

5. *If $E = \emptyset$, return failure. Otherwise, let $(j, v_j)$ be the last entry in $P$ such that $j \in E$. Remove $(j, v_j)$ from $P$ and, for each variable $k$ assigned a value after $j$, remove from $E_k$ any eliminating explanation that involves $j$. Set*

$$E_j = E_j \cup \epsilon(P, j) \cup \{(v_j, E \cap \overline{P})\} \qquad (2)$$

*so that $v_j$ is eliminated as a value for $j$ because of the values taken by variables in $E \cap \overline{P}$. The inclusion of the term $\epsilon(P, j)$ incorporates new information from variables that have been assigned values since the original assignment of $v_j$ to $j$. Now set $i = j$ and return to step 3.*

**Theorem 4.2** *Dynamic backtracking always terminates and is complete. It continues to satisfy Proposition 3.5 and can be expected to expand fewer nodes than backjumping provided that the goal nodes are distributed randomly in the search space.*

The essential difference between dynamic and dependency-directed backtracking is that the structure of our eliminating explanations means that we only save nogood information based on the current values of assigned variables; if a nogood depends on outdated information, we drop it. By doing this, we avoid the need to retain an exponential amount of nogood information. What makes this technique valuable is that (as stated in the theorem) termination is still guaranteed.

There is one trivial modification that we can make to Algorithm 4.1 that is quite useful in practice. After removing the current value for the backtrack variable $j$, Algorithm 4.1 immediately replaces it with another. But there is no real reason to do this; we could instead pick a value for an entirely different variable:

**Algorithm 4.3 (Dynamic backtracking)** *Given as inputs a constraint-satisfaction problem and an elimination mechanism $\epsilon$:*

1. *Set $P = E_i = \emptyset$ for each $i \in I$.*

2. *If $\overline{P} = I$, return $P$. Otherwise, select a variable $i \in I - \overline{P}$. Set $E_i = E_i \cup \epsilon(P, i)$.*

3. *Set $S = V_i - \widehat{E}_i$. If $S$ is nonempty, choose an element $v \in S$. Add $(i, v)$ to $P$ and return to step 2.*

4. *If $S$ is empty, we must have $\widehat{E}_i = V_i$; let $E$ be the set of all variables appearing in the explanations for each eliminated value.*

5. *If $E = \emptyset$, return failure. Otherwise, let $(j, v_j)$ be the last entry in $P$ that binds a variable appearing in $E$. Remove $(j, v_j)$ from $P$ and, for each variable $k$ assigned a value after $j$, remove from $E_k$ any eliminating explanation that involves $j$. Add $(v_j, E \cap \overline{P})$ to $E_j$ and return to step 2.*

## 5. An example

In order to make Algorithm 4.3 a bit clearer, suppose that we consider a small map-coloring problem in detail. The map is shown in Figure 1 and consists of five countries: Albania, Bulgaria, Czechoslovakia, Denmark and England. We will assume (wrongly!) that the countries border each other as shown in the figure, where countries are denoted by nodes and border one another if and only if there is an arc connecting them.

In coloring the map, we can use the three colors red, yellow and blue. We will typically abbreviate the country names to single letters in the obvious way.

We begin our search with Albania, deciding (say) to color it red. When we now look at Bulgaria, no colors are eliminated because Albania and Bulgaria do not share a border; we decide to color Bulgaria yellow. (This is a mistake.)

We now go on to consider Czechoslovakia; since it borders Albania, the color red is eliminated. We decide to color Czechoslovakia blue and the situation is now this:

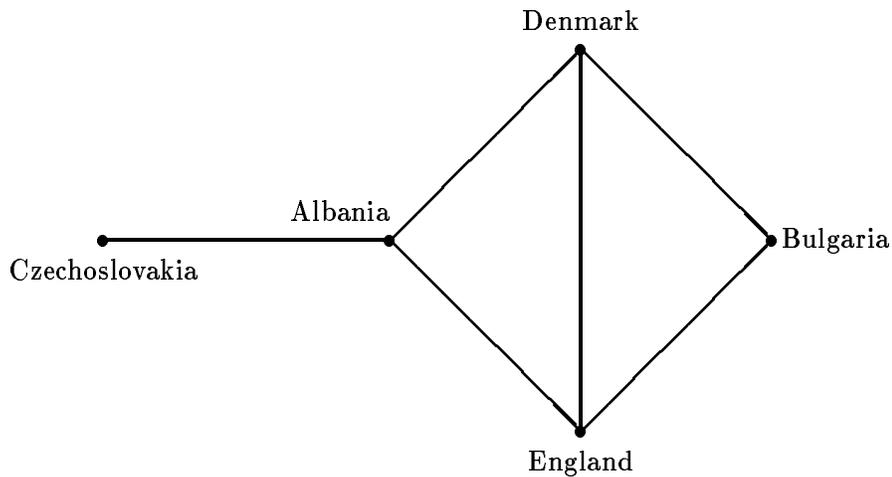

Figure 1: A small map-coloring problem

| country | color | red | yellow | blue |
|---|---|---|---|---|
| Albania | red | | | |
| Bulgaria | yellow | | | |
| Czechoslovakia | blue | A | | |
| Denmark | | | | |
| England | | | | |

For each country, we indicate its current color and the eliminating explanations that mean it cannot be colored each of the three colors (when such explanations exist). We now look at Denmark.

Denmark cannot be colored red because of its border with Albania and cannot be colored yellow because of its border with Bulgaria; it must therefore be colored blue. But now England cannot be colored any color at all because of its borders with Albania, Bulgaria and Denmark, and we therefore need to backtrack to one of these three countries. At this point, the elimination lists are as follows:

| country | color | red | yellow | blue |
|---|---|---|---|---|
| Albania | red | | | |
| Bulgaria | yellow | | | |
| Czechoslovakia | blue | A | | |
| Denmark | blue | A | B | |
| England | | A | B | D |

We backtrack to Denmark because it is the most recent of the three possibilities, and begin by removing any eliminating explanation involving Denmark from the above table to get:

| country | color | red | yellow | blue |
|---|---|---|---|---|
| Albania | red | | | |
| Bulgaria | yellow | | | |
| Czechoslovakia | blue | A | | |
| Denmark | | A | B | |
| England | | A | B | |

Next, we add to Denmark's elimination list the pair

$$(\text{blue}, \{A, B\})$$

This indicates correctly that because of the current colors for Albania and Bulgaria, Denmark cannot be colored blue (because of the subsequent dead end at England). Since every color is now eliminated, we must backtrack to a country in the set $\{A, B\}$. Changing Czechoslovakia's color won't help and we must deal with Bulgaria instead. The elimination lists are now:

| country | color | red | yellow | blue |
|---|---|---|---|---|
| Albania | red | | | |
| Bulgaria | | | | |
| Czechoslovakia | blue | A | | |
| Denmark | | A | B | A,B |
| England | | A | B | |

We remove the eliminating explanations involving Bulgaria and also add to Bulgaria's elimination list the pair

$$(\text{yellow}, A)$$

indicating correctly that Bulgaria cannot be colored yellow because of the current choice of color for Albania (red).

The situation is now:

| country | color | red | yellow | blue |
|---|---|---|---|---|
| Albania | red | | | |
| Czechoslovakia | blue | A | | |
| Bulgaria | | | A | |
| Denmark | | A | | |
| England | | A | | |

We have moved Bulgaria past Czechoslovakia to reflect the search reordering in the algorithm. We can now complete the problem by coloring Bulgaria red, Denmark either yellow or blue, and England the color not used for Denmark.

This example is almost trivially simple, of course; the thing to note is that when we changed the color for Bulgaria, we retained both the blue color for Czechoslovakia and the information indicating that none of Czechoslovakia, Denmark and England could be red. In more complex examples, this information may be very hard-won and retaining it may save us a great deal of subsequent search effort.

Another feature of this specific example (and of the example of the introduction as well) is that the computational benefits of dynamic backtracking are a consequence of the automatic realization that the problem splits into disjoint subproblems. Other authors have also discussed the idea of applying divide-and-conquer techniques to CSPs (Seidel, 1981; Zabih, 1990), but their methods suffer from the disadvantage that they constrain the order in which unassigned variables are assigned values, perhaps at odds with the common heuristic of assigning values first to those variables that are most tightly constrained. Dynamic backtracking can also be expected to be of use in situations where the problem in question does *not* split into two or more disjoint subproblems.[1]

## 6. Experimentation

Dynamic backtracking has been incorporated into the crossword-puzzle generation program described in (Ginsberg, Frank, Halpin, & Torrance, 1990), and leads to significant performance improvements in that restricted domain. More specifically, the method was tested on the problem of generating 19 puzzles of sizes ranging from $2 \times 2$ to $13 \times 13$; each puzzle was attempted 100 times using both dynamic backtracking and simple backjumping. The dictionary was shuffled between solution attempts and a maximum of 1000 backtracks were permitted before the program was deemed to have failed.

In both cases, the algorithms were extended to include iterative broadening (Ginsberg & Harvey, 1992), the cheapest-first heuristic and forward checking. Cheapest-first has also been called "most constrained first" and selects for instantiation that variable with the fewest number of remaining possibilities (i.e., that variable for which it is cheapest to enumerate the possible values (Smith & Genesereth, 1985)). Forward checking prunes the set of possibilities for crossing words whenever a new word is entered and constitutes our experimental choice of elimination mechanism: at any point, words for which there is no legal crossing word are eliminated. This ensures that no word will be entered into the crossword if the word has *no* potential crossing words at some point. The cheapest-first heuristic would identify the problem at the next step in the search, but forward checking reduces the number of backtracks substantially. The "least-constraining" heuristic (Ginsberg et al., 1990) was *not* used; this heuristic suggests that each word slot be filled with the word that minimally constrains the subsequent search. The heuristic was not used because it would invalidate the technique of shuffling the dictionary between solution attempts in order to gather useful statistics.

The table in Figure 2 indicates the number of successful solution attempts (out of 100) for each of the two methods on each of the 19 crossword frames. Dynamic backtracking is more successful in six cases and less successful in none.

With regard to the number of nodes expanded by the two methods, consider the data presented in Figure 3, where we graph the average number of backtracks needed by the two methods.[2] Although initially comparable, dynamic backtracking provides increasing computational savings as the problems become more difficult. A somewhat broader set of experiments is described in (Jonsson & Ginsberg, 1993) and leads to similar conclusions.

There are some examples in (Jonsson & Ginsberg, 1993) where dynamic backtracking leads to performance degradation, however; a typical case appears in Figure 4.[3] In this

---

1. I am indebted to David McAllester for these observations.
2. Only 17 points are shown because no point is plotted where backjumping was unable to solve the problem.
3. The worst performance degradation observed was a factor of approximately 4.

| Frame | Dynamic backtracking | Backjumping | Frame | Dynamic backtracking | Backjumping |
|-------|---------------------|-------------|-------|---------------------|-------------|
| 1 | 100 | 100 | 11 | 100 | 98 |
| 2 | 100 | 100 | 12 | 100 | 100 |
| 3 | 100 | 100 | 13 | 100 | 100 |
| 4 | 100 | 100 | 14 | 100 | 100 |
| 5 | 100 | 100 | 15 | 99 | 14 |
| 6 | 100 | 100 | 16 | 100 | 26 |
| 7 | 100 | 100 | 17 | 100 | 30 |
| 8 | 100 | 100 | 18 | 61 | 0 |
| 9 | 100 | 100 | 19 | 10 | 0 |
| 10 | 100 | 100 | | | |

Figure 2: Number of problems solved successfully

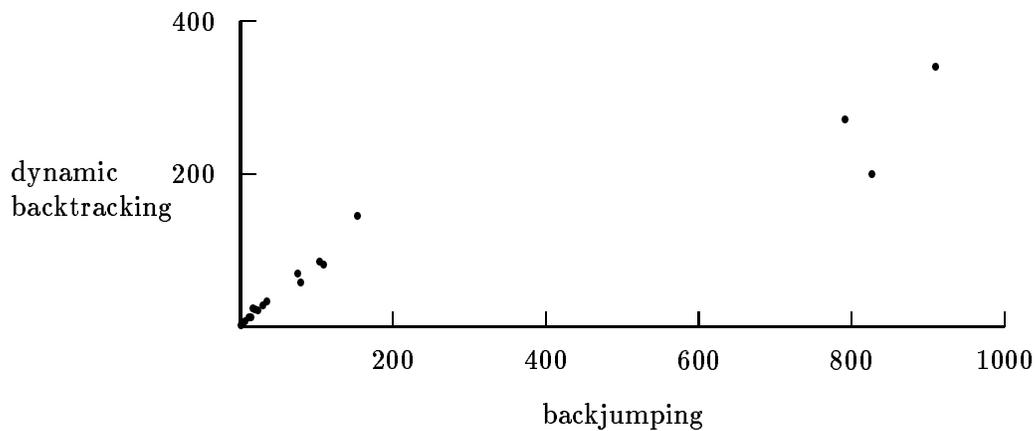

Figure 3: Number of backtracks needed

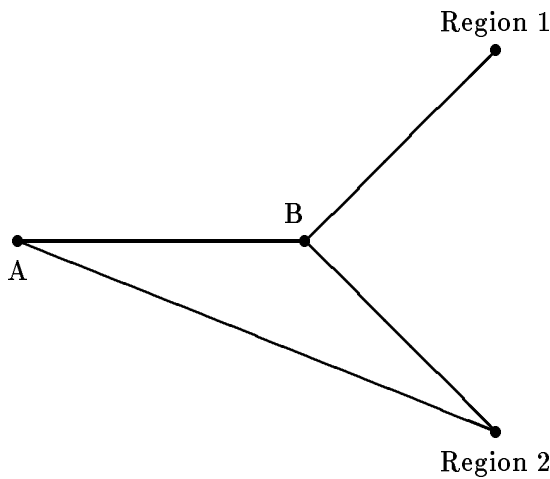

Figure 4: A difficult problem for dynamic backtracking

figure, we first color $A$, then $B$, then the countries in region 1, and then get stuck in region 2.

We now presumably backtrack directly to $B$, leaving the coloring of region 1 alone. But this may well be a mistake – the colors in region 1 will restrict our choices for $B$, perhaps making the subproblem consisting of $A$, $B$ and region 2 more difficult than it might be. If region 1 were easy to color, we would have been better off erasing it even though we didn't need to.

This analysis suggests that dependency-directed backtracking should also fare worse on those coloring problems where dynamic backtracking has trouble, and we are currently extending the experiments of (Jonsson & Ginsberg, 1993) to confirm this. If this conjecture is borne out, a variety of solutions come to mind. We might, for example, record how many backtracks are made to a node such as $B$ in the above figure, and then use this to determine that flexibility at $B$ is more important than retaining the choices made in region 1. The difficulty of finding a coloring for region 1 can also be determined from the number of backtracks involved in the search.

## 7. Summary

### 7.1 Why it works

There are two separate ideas that we have exploited in the development of Algorithm 4.3 and the others leading up to it. The first, and easily the most important, is the notion that it is possible to modify variable order on the fly in a way that allows us to retain the results of earlier work when backtracking to a variable that was assigned a value early in the search.

This reordering should not be confused with the work of authors who have suggested a dynamic choice among the variables that *remain* to be assigned values (Dechter & Meiri, 1989; Ginsberg et al., 1990; P. Purdom & Robertson, 1981; Zabih & McAllester, 1988); we are instead reordering the variables that have *been* assigned values in the search thus far.

Another way to look at this idea is that we have found a way to "erase" the value given to a variable directly as opposed to backtracking to it. This idea has also been explored by Minton *et.al.* in (Minton, Johnston, Philips, & Laird, 1990) and by Selman *et.al.* in (Selman, Levesque, & Mitchell, 1992); these authors also directly replace values assigned to variables in satisfiability problems. Unfortunately, the heuristic repair method used is incomplete because no dependency information is retained from one state of the problem solver to the next.

There is a third way to view this as well. The space that we are examining is really a graph, as opposed to a tree; we reach the same point by coloring Albania blue and then Bulgaria red as if we color them in the opposite order. When we decide to backjump from a particular node in the search space, we know that we need to back up until some particular property of that node ceases to hold – and the key idea is that by backtracking along a path *other than* the one by which the node was generated, we may be able to backtrack only slightly when we would otherwise need to retreat a great deal. This observation is interesting because it may well apply to problems other than CSPs. Unfortunately, it is not clear how to guarantee completeness for a search that discovers a node using one path and backtracks using another.

The other idea is less novel. As we have already remarked, our use of eliminating explanations is quite similar to the use of nogoods in the ATMS community; the principal difference is that we attach the explanations to the variables they impact and drop them when they cease to be relevant. (They might become relevant again later, of course.) This avoids the prohibitive space requirements of systems that permanently cache the results of their nogood calculations; this observation also may be extensible beyond the domain of CSPs specifically. Again, there are other ways to view this – Gashnig's notion of *backmarking* (Gaschnig, 1979) records similar information about the reason that particular portions of a search space are known not to contain solutions.

## 7.2 Future work

There are a variety of ways in which the techniques we have presented can be extended; in this section, we sketch a few of the more obvious ones.

### 7.2.1 BACKTRACKING TO OLDER CULPRITS

One extension to our work involves lifting the restriction in Algorithm 4.3 that the variable erased always be the most recently assigned member of the set $E$.

In general, we cannot do this while retaining the completeness of the search. Consider the following example:

Imagine that our CSP involves three variables, $x$, $y$ and $z$, that can each take the value 0 or 1. Further, suppose that this CSP has no solutions, in that after we pick any two values for $x$ and for $y$, we realize that there is no suitable choice for $z$.

We begin by taking $x = y = 0$; when we realize the need to backtrack, we introduce the nogood

$$x = 0 \supset y \neq 0 \qquad (3)$$

and replace the value for $y$ with $y = 1$.

This fails, too, but now suppose that we were to decide to backtrack to $x$, introducing the new nogood

$$y = 1 \supset x \neq 0 \qquad (4)$$

We change $x$'s value to 1 and erase (3).

This also fails. We decide that $y$ is the problem and change its value to 0, introducing the nogood

$$x = 1 \supset y \neq 1$$

but erasing (4). And when *this* fails, we are in danger of returning to $x = y = 0$, which we eliminated at the beginning of the example. This loop may cause a modified version of the dynamic backtracking algorithm to fail to terminate.

In terms of the proof of Theorem 4.2, the nogoods discovered already include information about all assigned variables, so there is no difference between (7) and (8). When we drop (3) in favor of (4), we are no longer in a position to recover (3).

We can deal with this by placing conditions on the variables to which we choose to backtrack; the conditions need to be defined so that the proof of Theorem 4.2 continues to hold.[4] Experimentation indicates that loops of the form we have described are extremely rare in practice; it may also be possible to detect them directly and thereby retain more substantial freedom in the choice of backtrack point.

This freedom of backtrack raises an important question that has not yet been addressed in the literature: When backtracking to avoid a difficulty of some sort, to where should one backtrack?

Previous work has been constrained to backtrack no further than the most recent choice that might impact the problem in question; any other decision would be both incomplete and inefficient. Although an extension of Algorithm 4.3 need not operate under this restriction, we have given no indication of how the backtrack point should be selected.

There are several easily identified factors that can be expected to bear on this choice. The first is that there remains a reason to expect backtracking to chronologically recent choices to be the most effective – these choices can be expected to have contributed to the fewest eliminating explanations, and there is obvious advantage to retaining as many eliminating explanations as possible from one point in the search to the next. It is possible, however, to simply identify that backtrack point that affects the fewest number of eliminating explanations and to use that.

Alternatively, it might be important to backtrack to the choice point for which there will be as many new choices as possible; as an extreme example, if there is a variable $i$ for which every value other than its current one has already been eliminated for other reasons, backtracking to $i$ is guaranteed to generate another backtrack immediately and should probably be avoided if possible.

---

4. Another solution appears in (McAllester, 1993).

Finally, there is some measure of the "directness" with which a variable bears on a problem. If we are unable to find a value for a particular variable $i$, it is probably sensible to backtrack to a second variable that shares a constraint with $i$ itself, as opposed to some variable that affects $i$ only indirectly.

How are these competing considerations to be weighed? I have no idea. But the framework we have developed is interesting because it allows us to work on this question. In more basic terms, we can now "debug" partial solutions to CSPs directly, moving laterally through the search space in an attempt to remain as close to a solution as possible. This sort of lateral movement seems central to human solution of difficult search problems, and it is encouraging to begin to understand it in a formal way.

### 7.2.2 DEPENDENCY PRUNING

It is often the case that when one value for a variable is eliminated while solving a CSP, others are eliminated as well. As an example, in solving a scheduling problem a particular choice of time (say $t = 16$) may be eliminated for a task $A$ because there then isn't enough time between $A$ and a subsequent task $B$; in this case, all later times can obviously be eliminated for $A$ as well.

Formalizing this can be subtle; after all, a later time for $A$ isn't *uniformly* worse than an earlier time because there may be other tasks that need to precede $A$ and making $A$ later makes that part of the schedule easier. It's the problem with $B$ alone that forces $A$ to be earlier; once again, the analysis depends on the ability to maintain dependency information as the search proceeds.

We can formalize this as follows. Given a CSP $(I, V, \kappa)$, suppose that the value $v$ has been assigned to some $i \in I$. Now we can construct a new CSP $(I', V', \kappa')$ involving the remaining variables $I' = I - \{i\}$, where the new set $V'$ need not mention the possible values $V_i$ for $i$, and where $\kappa'$ is generated from $\kappa$ by modifying the constraints to indicate that $i$ has been assigned the value $v$. We also make the following definition:

**Definition 7.1** *Given a* CSP, *suppose that $i$ is a variable that has two possible values $u$ and $v$. We will say that $v$ is* stricter *than $u$ if every constraint in the* CSP *induced by assigning $u$ to $i$ is also a constraint in the* CSP *induced by assigning $i$ the value $v$.*

The point, of course, is that if $v$ is stricter than $u$ is, there is no point to trying a solution involving $v$ once $u$ has been eliminated. After all, finding such a solution would involve satisfying all of the constraints in the $v$ restriction, these are a superset of those in the $u$ restriction, and we were unable to satisfy the constraints in the $u$ restriction originally.

The example with which we began this section now generalizes to the following:

**Proposition 7.2** *Suppose that a* CSP *involves a set $S$ of variables, and that we have a partial solution that assigns values to the variables in some subset $P \subseteq S$. Suppose further that if we extend this partial solution by assigning the value $u$ to a variable $i \notin P$, there is no further extension to a solution of the entire* CSP. *Now consider the* CSP *involving the variables in $S - P$ that is induced by the choices of values for variables in $P$. If $v$ is stricter than $u$ as a choice of value for $i$ in this problem, the original* CSP *has no solution that both assigns $v$ to $i$ and extends the given partial solution on $P$.* □

This proposition isn't quite enough; in the earlier example, the choice of $t = 17$ for $A$ will not be stricter than $t = 16$ if there is any task that needs to be scheduled before $A$ is. We need to record the fact that $B$ (which is no longer assigned a value) is the source of the difficulty. To do this, we need to augment the dependency information with which we are working.

More precisely, when we say that a set of variables $\{x_i\}$ eliminates a value $v$ for a variable $x$, we mean that our search to date has allowed us to conclude that

$$(v_1 = x_1) \wedge \cdots \wedge (v_k = x_k) \supset v \neq x$$

where the $v_i$ are the current choices for the $x_i$. We can obviously rewrite this as

$$(v_1 = x_1) \wedge \cdots \wedge (v_k = x_k) \wedge (v = x) \supset F \qquad (5)$$

where $F$ indicates that the CSP in question has no solution.

Let's be more specific still, indicating in (5) exactly *which* CSP has no solution:

$$(v_1 = x_1) \wedge \cdots \wedge (v_k = x_k) \wedge (v = x) \supset F(I) \qquad (6)$$

where $I$ is the set of variables in the complete CSP.

Now we can address the example with which we began this section; the CSP that is known to fail in an expression such as (6) is not the entire problem, but only a subset of it. In the example, we are considering, the subproblem involves only the two tasks $A$ and $B$. In general, we can augment our nogoods to include information about the subproblems on which they fail, and then measure strictness with respect to these restricted subproblems only. In our example, this will indeed allow us to eliminate $t = 17$ from consideration as a possible time for $A$.

The additional information stored with the nogoods doubles their size (we have to store a second subset of the variables in the CSP), and the variable sets involved can be manipulated easily as the search proceeds. The cost involved in employing this technique is therefore that of the strictness computation. This may be substantial given the data structures currently used to represent CSPs (which typically support the need to check if a constraint has been violated but little more), but it seems likely that compile-time modifications to these data structures can be used to make the strictness question easier to answer. In scheduling problems, preliminary experimental work shows that the idea is an important one; here, too, there is much to be done.

The basic lesson of dynamic backtracking is that by retaining only those nogoods that are still relevant given the partial solution with which we are working, the storage difficulties encountered by full dependency-directed methods can be alleviated. This is what makes all of the ideas we have proposed possible – erasing values, selecting alternate backtrack points, and dependency pruning. There are surely many other effective uses for a practical dependency maintenance system as well.

## Acknowledgements

This work has been supported by the Air Force Office of Scientific Research under grant number 92-0693 and by DARPA/Rome Labs under grant number F30602-91-C-0036. I would like to thank Rina Dechter, Mark Fox, Don Geddis, Will Harvey, Vipin Kumar, Scott Roy and Narinder Singh for helpful comments on these ideas. Ari Jonsson and David McAllester provided me invaluable assistance with the experimentation and proofs respectively.

## A. Proofs

**Lemma 2.4** *Let $\epsilon$ be a complete elimination mechanism for a* CSP, *let $P$ be a partial solution to this* CSP *and let $i \notin \overline{P}$. Now if $P$ can be successfully extended to a complete solution after assigning $i$ the value $v$, then $v \notin \widehat{\epsilon}(P, i)$.*

**Proof.** Suppose otherwise, so that $(v, E) \in \epsilon(P, i)$. It follows directly from the completeness of $\epsilon$ that

$$E \cap (\overline{P} - \overline{P}) \neq \emptyset$$

a contradiction. □

**Lemma 2.6** *At any point in the execution of Algorithm 2.5, if the last element of the partial solution $P$ assigns a value to the variable $i$, then the unexplored siblings of the current node are those that assign to $i$ the values in $V_i - E_i$.*

**Proof.** We first note that when we decide to assign a value to a new variable $i$ in step 2 of the algorithm, we take $E_i = \widehat{\epsilon}(P, i)$ so that $V_i - E_i$ is the set of allowed values for this variable. The lemma therefore holds in this case. The fact that it continues to hold through each repetition of the loop in steps 3 and 4 is now a simple induction; at each point, we add to $E_i$ the node that has just failed as a possible value to be assigned to $i$. □

**Proposition 2.7** *Algorithm 2.5 is equivalent to depth-first search and therefore complete.*

**Proof.** This is an easy consequence of the lemma. Partial solutions correspond to nodes in the search space. □

**Lemma 3.2** *Let $P$ be a partial solution obtained during the execution of Algorithm 3.1, and let $i \in \overline{P}$ be a variable assigned a value by $P$. Now if $P' \subseteq P$ can be successfully extended to a complete solution after assigning $i$ the value $v$ but $(v, E) \in E_i$, we must have*

$$E \cap (\overline{P} - \overline{P'}) \neq \emptyset$$

**Proof.** As in the proof of Lemma 2.6, we show that no step of Algorithm 3.1 can cause Lemma 3.2 to become false.

That the lemma holds after step 2, where the search is extended to consider a new variable, is an immediate consequence of the assumption that the elimination mechanism is complete.

In step 4, when we add $(v_j, E - \{j\})$ to the set of eliminating explanations for $j$, we are simply recording the fact that the search for a solution with $j$ set to $v_j$ failed because we were unable to extend the solution to $i$. It is a consequence of the inductive hypothesis that as long as no variable in $E - \{j\}$ changes, this conclusion will remain valid. □

**Proposition 3.4** *Backjumping is complete and always expands fewer nodes than does depth-first search.*

**Proof.** That fewer nodes are examined is clear; for completeness, it follows from Lemma 3.2 that the backtrack to some element of $E$ in step 5 will always be necessary if a solution is to be found. □

**Proposition 3.5** *The amount of space needed by backjumping is $o(i^2v)$, where $i = |I|$ is the number of variables in the problem and $v$ is the number of values for that variable with the largest value set $V_i$.*

**Proof.** The amount of space needed is dominated by the storage requirements of the elimination sets $E_j$; there are $i$ of these. Each one might refer to each of the possible values for a particular variable $j$; the space needed to store the reason that the value $j$ is eliminated is at most $|I|$, since the reason is simply a list of variables that have been assigned values. There will never be two eliminating explanations for the same variable, since $\epsilon$ is concise and we never rebind a variable to a value that has been eliminated. □

**Theorem 4.2** *Dynamic backtracking always terminates and is complete. It continues to satisfy Proposition 3.5 and can be expected to expand fewer nodes than backjumping provided that the goal nodes are distributed randomly in the search space.*

**Proof.** There are four things we need to show: That dynamic backtracking needs $o(i^2v)$ space, that it is complete, that it can be expected to expand fewer nodes than backjumping, and that it terminates. We prove things in this order.

**Space** This is clear; the amount of space needed continues to be bounded by the structure of the eliminating explanations.

**Completeness** This is also clear, since by Lemma 3.2, all of the eliminating explanations retained in the algorithm are obviously still valid. The new explanations added in (2) are also obviously correct, since they indicate that $j$ cannot take the value $v_j$ as in backjumping and that $j$ also cannot take any values that are eliminated by the variables being backjumped over.

**Efficiency** To see that we *expect* to expand fewer nodes, suppose that the subproblem involving only the variables being jumped over has $s$ solutions in total, one of which is given by the existing variable assignments. Assuming that the solutions are distributed randomly in the search space, there is at least a $1/s$ chance that this particular solution leads to a solution of the entire CSP; if so, the reordered search – which considers this solution earlier than the other – will save the expense of either assigning new values to these variables or repeating the search that led to the existing choices. The reordered search will also benefit from the information in the nogoods that have been retained for the variables being jumped over.

**Termination** This is the most difficult part of the proof.

As we work through the algorithm, we will be generating (and then discarding) a variety of eliminating explanations. Suppose that $e$ is such an explanation, saying that $j$ cannot take the value $v_j$ because of the values currently taken by the variables in some set $e_V$. We will denote the variables in $e_V$ by $x_1, \ldots, x_k$ and their current values by $v_1, \ldots, v_k$. In declarative terms, the eliminating explanation is telling us that

$$(x_1 = v_1) \wedge \cdots \wedge (x_k = v_k) \supset j \neq v_j \tag{7}$$

Dependency-directed backtracking would have us accumulate all of these nogoods; dynamic backtracking allows us to drop any particular instance of (7) for which the antecedent is no longer valid.

The reason that *dependency-directed* backtracking is guaranteed to terminate is that the set of accumulated nogoods eliminates a monotonically increasing amount of the search space. Each nogood eliminates a new section of the search space because the nature of the search process is such that any node examined is consistent with the nogoods that have been accumulated thus far; the process is monotonic because all nogoods are retained throughout the search. These arguments cannot be applied to dynamic backtracking, since nogoods are forgotten as the search proceeds. But we can make an analogous argument.

To do this, suppose that when we discover a nogood like (7), we record with it all of the variables that precede the variable $j$ in the partial order, together with the values currently assigned to these variables. Thus an eliminating explanation becomes essentially a nogood $n$ of the form (7) together with a set $S$ of variable/value pairs.

We now define a mapping $\lambda(n, S)$ that changes the antecedent of (7) to include assumptions about *all* the variables bound in $S$, so that if $S = \{s_i, v_i\}$,

$$\lambda(n, S) = [(s_1 = v_1) \wedge \cdots \wedge (s_l = v_l) \supset j \neq v_j] \tag{8}$$

At any point in the execution of the algorithm, we denote by $N$ the conjunction of the modified nogoods of the form (8).

We now make the following claims:

1. For any eliminating explanation $(n, S)$, $n \models \lambda(n, S)$ so that $\lambda(n, S)$ is valid for the problem at hand.

2. For any new eliminating explanation $(n, S)$, $\lambda(n, S)$ is not a consequence of $N$.

3. The deductive consequences of $N$ grow monotonically as the dynamic backtracking algorithm proceeds.

The theorem will follow from these three observations, since we will know that $N$ is a valid set of conclusions for our search problem and that we are once again making monotonic progress toward eliminating the entire search space and concluding that the problem is unsolvable.

That $\lambda(n, S)$ is a consequence of $(n, S)$ is clear, since the modification used to obtain (8) from (7) involves strengthening that antecedent of (7). It is also clear that $\lambda(n, S)$ is not a consequence of the nogoods already obtained, since we have added to the antecedent only conditions that hold for the node of the search space currently under examination. If $\lambda(n, S)$ were a consequence of the nogoods we had obtained thus far, this node would not be being considered.

The last observation depends on the following lemma:

**Lemma A.1** *Suppose that $x$ is a variable assigned a value by our partial solution and that $x$ appears in the antecedent of the nogood $n$ in the pair $(n, S)$. Then if $S'$ is the set of variables assigned values no later than $x$, $S' \subseteq S$.*

**Proof.** Consider a $y \in S'$, and suppose that it were not in $S$. We cannot have $y = x$, since $y$ would then be mentioned in the nogood $n$ and therefore in $S$. So we can suppose that $y$ is actually assigned a value *earlier* than $x$ is. Now when $(n, S)$ was added to the set of eliminating explanations, it must have been the case that $x$ was assigned a value (since it appears in the antecedent of $n$) but that $y$ was not. But we also know that there was a later time when $y$ was assigned a value but $x$ was not, since $y$ precedes $x$ in the current partial solution. This means that $x$ must have changed value at some point after $(n, S)$ was added to the set of eliminating explanations – but $(n, S)$ would have been deleted when this happened. This contradiction completes the proof.  □

Returning to the proof the Theorem 4.2, suppose that we eventually drop $(n, S)$ from our collection of nogoods and that when we do so, the new nogood being added is $(n', S')$. It follows from the lemma that $S' \subseteq S$. Since $x_i = v_i$ is a clause in the antecedent of $\lambda(n, S)$, it follows that $\lambda(n', S')$ will imply the negation of the antecedent of $\lambda(n, S)$ and will therefore imply $\lambda(n, S)$ itself. Although we drop $\lambda(n, S)$ when we drop the nogood $(n, S)$, $\lambda(n, S)$ continues to be entailed by the modified set $N$, the consequences of which are seen to be growing monotonically.  □